**Titre :** Lumière et 3D : une exploration méthodologique des techniques de numérisation adaptées à une sélection d'objets du musée d'Archéologie nationale.

**Résumé :** La nécessité de numériser les objets du patrimoine fait aujourd'hui consensus. Cet article présente le contexte très en vogue de la création de « jumeaux numériques ». Il illustre la diversité des méthodes de numérisation 3D *photographique*, mais là n'est pas son seul objectif. Grâce à une sélection d'objets provenant des collections du musée d'Archéologie nationale, il montre qu'aucune méthode ne répond à tous les cas de figure. La méthode à préconiser pour tel ou tel objet doit plutôt résulter d'un choix concerté entre les acteurs du patrimoine et ceux du domaine du numérique, chaque nouvel objet pouvant nécessiter d'adapter les outils existants. Il est donc vain de chercher à réaliser une classification *absolue* des méthodes de numérisation 3D. A contrario, il s'agit de trouver pour chaque objet d'étude l'outil numérique approprié, en prenant en compte non seulement ses caractéristiques, mais également l'usage futur de son double numérique.

**Abstract :** The need to digitize heritage objects is now widely accepted. This article presents the very fashionable context of the creation of "digital twins". It illustrates the diversity of photographic 3D digitization methods, but this is not its only objective. Using a selection of objects from the collections of the musée d'Archéologie nationale, it shows that no single method is suitable for all cases. Rather, the method to be recommended for a given object should be the result of a concerted choice between those involved in heritage and those involved in the digital domain, as each new object may require the adaptation of existing tools. It would therefore be pointless to attempt an absolute classification of 3D digitization methods. On the contrary, we need to find the digital tool best suited to each object, taking into account not only its characteristics, but also the future use of its digital twin.

**Auteurs**

- Antoine Laurent, doctorant, Institut de recherche en informatique de Toulouse (IRIT, UMR CNRS 5505) et Laboratoire d'archéologie de Toulouse (TRACES, UMR CNRS 5608),
- Jean Mélou, post-doctorant, Institut national polytechnique de Toulouse (INPT) et Institut de recherche en informatique de Toulouse (IRIT),
- Catherine Schwab, conservatrice en chef du patrimoine, responsable des collections paléolithiques et mésolithiques, musée d'Archéologie nationale, Saint-Germain-en-Laye,
- Rolande Simon-Millot, conservatrice en chef du patrimoine, responsable des collections néolithique et âge du Bronze, musée d'Archéologie nationale, Saint-Germain-en-Laye,
- Sophie Feret, conservatrice en chef du patrimoine, responsable des collections de Gaule romaine, musée d'Archéologie nationale, Saint-Germain-en-Laye,
- Thomas Sagory, contractuel, responsable du développement numérique, musée d'Archéologie nationale, Saint-Germain-en-Laye,

- Carole Fritz, directrice de recherche CNRS, Laboratoire d'archéologie moléculaire et structurale (LAMS, Paris, UMR CNRS 8220),
- Jean-Denis Durou, enseignant-chercheur, Institut national polytechnique de Toulouse (INPT) et Institut de recherche en informatique de Toulouse (IRIT).

**Introduction**

Ce travail collectif s'inscrit dans le cadre de la thèse de doctorat d'Antoine Laurent[1], qui vise à trouver des points de convergence entre archéologues et chercheurs en vision 3D, ces deux communautés pouvant avoir des regards croisés, voire antagonistes, sur la numérisation. Le concepteur d'une méthode de numérisation 3D choisit généralement les objets les plus à même de la mettre en valeur, à l'inverse des besoins des acteurs du patrimoine, qui doivent trouver, pour chaque objet, la « meilleure » solution de numérisation. Or, ces besoins eux-mêmes peuvent être extrêmement variés, selon que le but visé est, par exemple, l'archivage ou la consultation à distance.

Il nous a semblé opportun d'inscrire cette exploration méthodologique des techniques de numérisation 3D dans l'environnement d'une structure muséale, en l'occurrence le musée d'Archéologie nationale de Saint-Germain-en-Laye, en faisant appel à un investissement matériel limité et à une suite logicielle *open-source* pour le traitement des données.

**Le développement numérique au musée d'Archéologie nationale, domaine national de Saint-Germain-en-Laye**

L'atelier de prise de vues du service du développement numérique du musée d'Archéologie nationale mène régulièrement des campagnes de numérisation des objets des collections du MAN afin d'étudier, de documenter et de diffuser des données de qualité[2]. Si la photogrammétrie est actuellement la méthode la plus couramment utilisée au musée, selon les besoins et les problématiques, d'autres méthodes peuvent être exploitées. Le musée est associé à différents projets de recherche et développement, comme cela est le cas avec l'IRIT (Institut de recherche en informatique de Toulouse) dans le cadre du projet doctoral *Archaeoroom* d'Antoine Laurent.

Pour le MAN, l'objectif est d'être en mesure d'assurer une production de données maîtrisées, exploitable et librement diffusable. Cette démarche s'inscrit dans une longue tradition, puisque dès sa création le musée disposait d'un atelier de moulages[3], dont les objectifs sont assez proches de ceux de l'atelier de prise de vues, à la différence près que certains objets composites trop fragiles pour être moulés peuvent aujourd'hui être numérisés sans contact. En 2023, le projet scientifique et culturel du MAN[4] confirme cet engagement et la volonté de concevoir un jumeau numérique s'appuyant sur une infrastructure repensée et sécurisée, sur la mise en ligne des collections (image et 3D), sur la création d'un portail agrégatif, voire d'un métavers, sur un espace immersif ainsi que sur la structuration de partenariats avec les acteurs du numérique.

---

[1] La thèse ArchaeoRoom a pour sujet « la 3D au service du patrimoine archéologique pour son étude et sa conservation dans la science ouverte ». Elle s'effectue sous la direction de Jean-Denis Durou et de Carole Fritz, et est financée à parts égales par la région Occitanie et l'université fédérale de Toulouse (programme 2021 des allocations doctorales interdisciplinaires). Le sujet interdisciplinaire articule la création de modèles numériques par reconstruction 3D photographique et interroge leur utilité pour les sciences du patrimoine, notamment pour l'analyse surfacique (géométrie et couleur).
[2] Sagory 2018.
[3] Douau 1984.

**Enregistrer le relief d'objets patrimoniaux**

Pour commencer, il nous semble nécessaire de balayer deux idées reçues. Primo, de par la complexité de la numérisation 3D, aucune méthode ne permet à elle seule de restituer toutes les composantes d'un objet archéologique. Secundo, l'usage du jumeau numérique ne peut en aucun cas se substituer à l'objet lui-même. Il sera donc toujours nécessaire, pour certaines études, de revenir à l'objet réel.

Pour répondre aux besoins en matière d'étude des œuvres, les premières méthodes utilisées ont été directes. Il se produisait alors un contact avec l'objet : dans le cas des statues-menhirs, par le biais de moulages, de calques à même la surface ou encore par « frottage ». L'intérêt était d'être au plus près de l'œuvre en limitant les déformations pour le relevé, tout en conservant l'échelle 1. La pratique de ces techniques a été progressivement abandonnée, afin de ne pas endommager la surface[5]. Les techniques indirectes comme la photographie, le croquis ou encore les dessins aux points, qui ont pris le relais, avant de devenir numériques, répondent à des critères de mise en forme afin de permettre une étude interprétative de l'objet[6]. Ces normes sont généralement les suivantes : échelle, légende, charte colorimétrique, sémiologie ou encore mise en lumière particulière.

Les critères à prendre en compte pour choisir une méthode de numérisation 3D sont multiples. Comme nous l'avons déjà dit, ce choix dépend non seulement du type d'objet à numériser, mais aussi de l'usage qui sera fait du « jumeau numérique »[7]. Les acteurs du patrimoine n'ont le plus souvent accès qu'à l'information de surface d'un objet physique, notamment à cause des technologies actuelles qui reconstruisent mieux les objets opaques et mats que les matériaux brillants et/ou translucides. Cependant il est appréciable de numériser l'intérieur des objets, par exemple l'intérieur de la matière pour lire les fibres du bois, ou encore d'accéder à une surface métallique prise dans une concrétion. Les méthodes de numérisation volumique telles que la tomographie, connue du grand public sous le terme de « scanner », qui sont couramment utilisées dans le domaine médical, le sont de plus en plus par les acteurs du patrimoine grâce à la mise en commun de matériel dédié[8].

Les quatre techniques surfaciques dominantes dans le domaine du patrimoine sont la photogrammétrie, la lasergrammétrie, les scanners à SLC ou lumière structurée et la RTI. Dans cet article, nous nous focalisons sur les techniques de numérisation 3D photographique, qui opèrent dans le domaine visible et sont faciles à mettre en œuvre avec du matériel grand public, et dont les données peuvent être exploitées en *open-source*. Ces techniques sont la photogrammétrie et la stéréophotométrie, dont le principe est analogue à celui de la RTI.

---

[4] Mousseaux 2023. Le PSC de 2017 est consultable : https://musee-archeologienationale.fr/projet-scientifique-et-culturel
[5] Maillé 2010.
[6] Djindjian 2011.
[7] Niccollucci 2023.
[8] Nous pouvons souligner ici le travail du C2RMF.

La reconstruction 3D photographique consiste à combiner plusieurs photographies d'un objet pour produire un modèle tridimensionnel de sa surface. Il n'est pas absolument nécessaire de prendre plusieurs clichés d'un objet pour obtenir sa représentation 3D. Les instruments connus sous le nom de « capteurs de profondeur » (la fameuse Kinect) fournissent une cartographie 3D de l'environnement en temps réel, mais ces outils, spécialement conçus pour les jeux vidéo, n'atteignent pas la précision requise dans le domaine patrimonial. Il en va de même des méthodes reposant sur le principe très en vogue de l'apprentissage profond (*deep learning*), qui s'inspirent des capacités d'interprétation de notre système visuel. Ayant appris à associer un relief à chaque image d'un très grand nombre d'objets, un réseau de neurones artificiels suffisamment entraîné infère la géométrie d'un objet à partir d'une seule photographie. Mais là encore, la précision n'est pas suffisante, par exemple pour l'étude des gravures fines. Pour ce faire, il est indispensable d'utiliser plusieurs photographies[9].

Quels que soient les outils choisis, deux critères importants sont à prendre en compte : la résolution, qui désigne l'espacement entre deux informations, et la précision, qui désigne la fidélité du relevé 3D à l'objet réel en abscisse, ordonnée et profondeur. Mais, alors que la résolution est déterminée dès la captation des données, la précision du modèle 3D dépend fortement des traitements qui leur sont appliqués. Or, le choix d'un niveau de résolution et/ou de précision découle aussi de l'usage que l'on souhaite faire de ces données, ce qui détermine l'échelle de travail. Il est notable que ces paramètres sont rarement précisés dans un cahier des charges[10]. Parmi les exemples développés dans cet article, le niveau de besoin est centimétrique pour la statuaire néolithique, car l'objectif est une valorisation en ligne avec des modèles allégés. En revanche, pour l'étude d'une gravure fine sur un coquillage du Paléolithique, il est nécessaire d'atteindre une résolution de quelques dizaines de microns.

Il existe plusieurs façons de visualiser un jumeau numérique, d'y accéder ou de l'interroger, qui peuvent dépendre de la méthode employée. Bien sûr, suite à une reconstruction 3D, les représentations les plus courantes sont les nuages de points colorés et les maillages texturés photoréalistes. Le second mode de visualisation donne à voir une surface continue sur laquelle peut être plaquée une image ou une texture, là où le premier mode est plus rapide à afficher, qui plus est avec une meilleure précision[11]. Comme il n'est pas toujours nécessaire d'accéder à l'information sur la troisième dimension, d'autres types de représentation 2D existent : les images géoréférencées (par exemple, par la photogrammétrie), les champs de normales RVB (rouge, vert, bleu) qui informent sur l'orientation locale du relief, ou encore les cartes de profondeur qui apportent une lecture de la distance d'une surface vue sous un angle donné. Ces modèles peuvent être augmentés par des données multispectrales et par l'imagerie UV afin de sortir du spectre visible. Pour aller plus loin, les modèles 3D seront complétés par des annotations sémantiques et par les observations de chaque spécialiste qui gravite autour de l'œuvre[12]. Ce dernier point interroge les problématiques liées à la représentation de ces superpositions d'informations[13].

---

[9] Hernandez 2004.
[10] Granier, Chayani, Abergel, *et al.* 2019.
[11] Granier, Chayani, Abergel, *et al.* 2019.
[12] Talon, Cauvin-Hardy, Chateauneuf 2017.
[13] De Luca, Abergel, Guillem, *et al.* 2022.

Le recours à la numérisation 3D d'un objet peut donner lieu à une gradation dans les résultats possibles ou attendus : cela va du résultat idéal, à savoir un modèle 3D de l'objet très précis et sa réflectance[14] (également appelée BRDF), à un résultat plus modeste qui peut quand même répondre aux besoins concernant l'étude, la conservation, le suivi de l'usure[15], la restauration, la visite virtuelle, la synthèse de vues, etc.

Certaines méthodes de numérisation 3D paraissent performantes, alors que c'est parfois l'interprétation du rendu par le cerveau qui nous donne l'illusion que l'objet virtuel est proche de l'objet original. En effet, le système visuel humain est facilement trompé par le plaquage d'une texture réaliste sur un relief grossier. Cela peut suffire, par exemple, pour visualiser l'objet à distance, mais en aucun cas pour effectuer des mesures de précision. Plus généralement, il faut bien distinguer les données brutes, les données traitées et les données interprétées (qui résultent d'une étude par un expert d'un domaine).

Enfin, dans le respect des artéfacts et pour leur préservation, il convient de s'adapter aux contraintes des objets, du lieu, des outils, du temps de captation des données et du coût de telle ou telle méthode de numérisation 3D. Par exemple, il est souvent plus simple de déplacer le matériel de numérisation plutôt que l'objet à numériser.

**Numérisation 3D « classique » d'une statue-menhir en pierre par photogrammétrie**

Un objet mat suffisamment texturé, comme la pierre brute, se prête bien à la numérisation 3D par photogrammétrie. En effet, cette technique repose sur le principe de la « triangulation », qui nécessite de savoir repérer un même point 3D vu sous différents angles. Elle est d'autant plus appropriée lorsque l'objet requiert la lecture de chaque face (avant / arrière / côtés / haut / bas). Le MAN dispose du matériel (un simple appareil photographique suffit) et des compétences requises pour la mise en œuvre de cette méthode de numérisation 3D. L'objet choisi pour l'illustrer ici est une statue-menhir.

À la fin du 4$^e$ millénaire, différents groupes culturels pratiquant l'agriculture, l'élevage et la métallurgie du cuivre occupent le sud de la France[16]. Certains d'entre eux développent une statuaire très particulière qui met en exergue la figure humaine, sculptée en ronde-bosse sur des blocs monolithes grossièrement aménagés et fichés en terre[17]. Les premiers érudits du XIX$^e$ siècle, à la vue de ces pierres singulières, les ont nommées « statues-menhirs ». L'abbé Frédéric Hermet (1856-1939), curé de l'Hospitalet dans l'Aveyron, est l'auteur de plusieurs articles écrits entre 1888 et 1898 consacrés à ces « divinités archaïques »[18]. Ses découvertes suscitent l'intérêt de nombreux chercheurs, si bien que quatre d'entre elles sont envoyées à Paris en 1900, pour être présentées lors de l'Exposition universelle du Trocadéro[19].

À l'issue de l'exposition, le musée de Saint-Germain-en-Laye les acquiert au prix de 100 francs pièce. C'est ainsi que quatre statues-menhirs, trois de l'Aveyron et une du Tarn, firent leur apparition au musée. L'unique statue-menhir complètement masculine de l'ensemble (MAN 46046) a été découverte en 1887 au sommet d'une montagne appelée Puech-Réal. D'une hauteur de 85 cm, elle porte distinctement un baudrier et un objet au niveau des mains. Ce « motif » est difficile à identifier avec certitude. On s'accorde généralement à reconnaître son caractère emblématique, en lien probable avec la représentation du pouvoir et de l'autorité. Il semble que cette statue ait servi de pierre à aiguiser pendant quelques années après sa découverte, car une partie de son visage a été manifestement lissée par abrasion.

---

[14] La réflectance est la grandeur photométrique qui décrit comment une surface réémet la lumière reçue.
[15] Schwab 2021.
[16] Arnal 1976.
[17] D'anna 1977.
[18] Hermet 1898, 1900.
[19] Hermet 1908.

La figure 1 illustre la séance de prise de vues de cette statue-menhir dans l'espace de visite alors en cours de déménagement des collections. Afin d'obtenir un recouvrement suffisant entre les points de vue, ce sont près de deux cents photographies qui ont été nécessaires. L'utilisation d'un éclairage uniforme fixe, réalisé à l'aide de plusieurs panneaux de LEDs, permet d'éviter les artéfacts qui pourraient résulter de correspondances erronées entre vues.

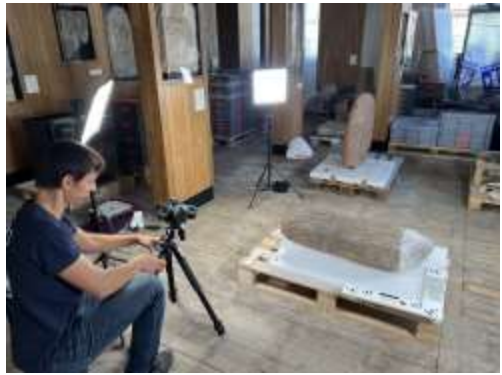

Figure 1 - La photogrammétrie, appliquée à une centaine de vues par face de la statue-menhir de Puech-Réal, prises sous différentes poses, sous un éclairage uniforme réalisé à l'aide de panneaux de LEDs, fournit le résultat de la figure 2.

La figure 2 montre le modèle 3D obtenu par photogrammétrie. Le rééclairage sous plusieurs directions du modèle 3D non coloré, qui met en évidence différentes caractéristiques du relief, constitue une aide précieuse à la lecture. En revanche, la couleur provient des images elles-mêmes, au risque d'insérer des ombres dans le modèle 3D coloré. La technique appropriée pour obtenir la couleur intrinsèque de l'objet s'appelle la stéréophotométrie[20].

---

[20] Laurent 2022.

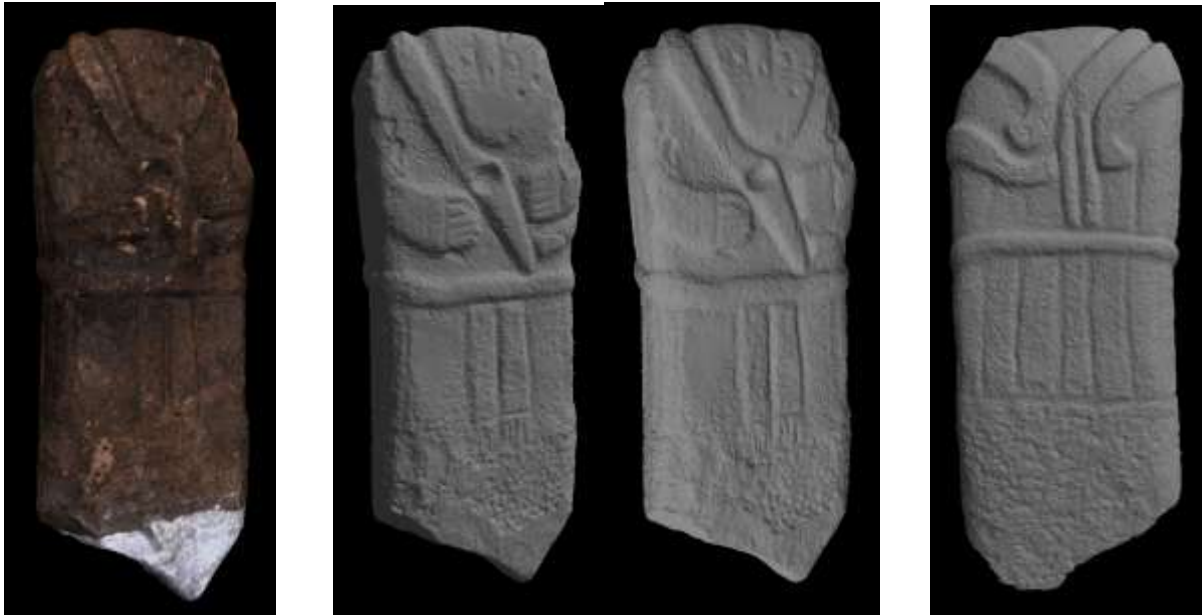

Figure 2 - Résultat obtenu par photogrammétrie, à partir de deux cents vues de la statue-menhir de Puech-Réal. De gauche à droite : modèle 3D coloré ; vue de la face avant sous deux éclairages différents ; vue de la face arrière.

**Numérisation 3D par stéréophotométrie : comparaison avec la photogrammétrie**

La photogrammétrie, qui est la méthode de numérisation 3D la plus répandue dans le domaine du patrimoine, est inopérante pour les objets lisses sans texture. Elle fournit alors des modèles 3D imprécis, voire très éloignés de la réalité. Une méthode adaptée à de tels objets est la stéréophotométrie, qui consiste à prendre plusieurs clichés de l'objet sous le même angle, en faisant seulement varier l'éclairage[21]. Un autre avantage de cette méthode est la résolution du modèle 3D produit, qui est égale à celle du capteur, car un même pixel correspond au même point 3D dans toutes les images. Un troisième avantage très appréciable de la stéréophotométrie est qu'elle fournit, en plus du relief, la couleur intrinsèque de l'objet photographié (dénuée de l'ombrage), par opposition à sa couleur apparente, ce qui présente un avantage décisif pour les études de collections[22] ou pour les applications de réalité augmentée. Enfin, l'acquisition et le traitement des données sont beaucoup plus rapides par stéréophotométrie que par photogrammétrie.

Comme nous l'avons dit en introduction, aucune méthode, aussi séduisante soit-elle, ne concentre tous les avantages. Contrairement à la photogrammétrie, la stéréophotométrie ne peut être mise en œuvre avec un simple *smartphone*, ce qui explique sans doute qu'elle soit inconnue du grand public, car elle nécessite de contrôler l'éclairage. Mais s'il est effectivement difficile de l'utiliser en extérieur, cela n'est pas un obstacle dans les musées. Par ailleurs, dans la mesure où la pose de l'appareil photographique est fixe, seule une partie de la surface est reconstruite. Nous avons donc choisi, pour illustrer cette méthode, plusieurs objets dont une des faces suffit à constituer un sujet d'étude. Enfin, il faut non seulement contrôler la lumière, mais également savoir l'estimer.

---

[21] Durou 2020.
[22] Il est intéressant pour un scientifique d'être assuré que la couleur de rendu du modèle 3D est fidèle à l'original.

**Mise en évidence par stéréophotométrie des détails fins d'une tablette en ivoire**

En 1967 puis en 1968, à 8 m de profondeur, au fond d'un puits, Jean-Paul Bertaux[23] découvre plusieurs centaines de fragments de tablettes en ivoire d'éléphant formant deux diptyques, mesurant chacun 29 cm sur 19 cm, datés du II$^e$ siècle. L'un est conservé au musée départemental d'Art ancien et contemporain à Épinal (M0536_2013.0.217) et l'autre au MAN (MAN 83675), à la demande d'André Malraux, alors ministre des Affaires culturelles. Les décors représentent une carte du ciel et un zodiaque sont gravés au ciseau et à la gouge. Le tout était rehaussé de couleurs (rouge vermillon, noir de galène, jaune d'orpiment et feuilles d'or) qui ont été étudiées récemment par le Centre de recherche et de restauration des musées de France (C2RMF)[24].

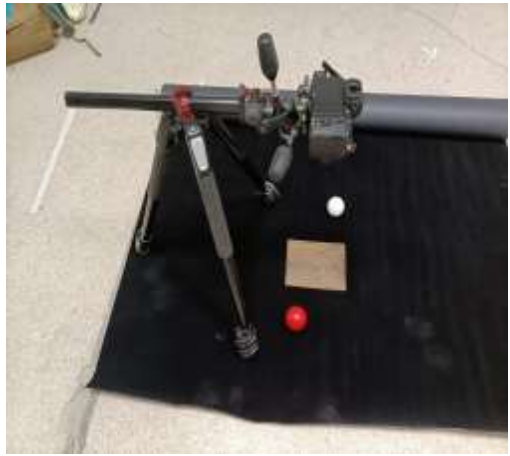

Figure 3 - La stéréophotométrie nécessite de poser l'appareil photographique sur un pied, afin qu'un même pixel corresponde au même point 3D dans les différentes images. Les sphères permettent d'estimer la direction et l'intensité de chaque éclairage.

Le montage illustré sur la figure 3 n'est guère plus sophistiqué que celui de la photogrammétrie, si ce n'est que les éclairages doivent être directionnels. D'autre part, il est nécessaire de disposer une ou plusieurs sphères de couleur unie au voisinage de l'objet, afin d'estimer la direction et l'intensité de chaque éclairage, en veillant à ce que cela ne gêne pas les acquisitions. La figure 4 montre deux images, parmi une dizaine, de la tablette zodiacale sous deux éclairages différents.

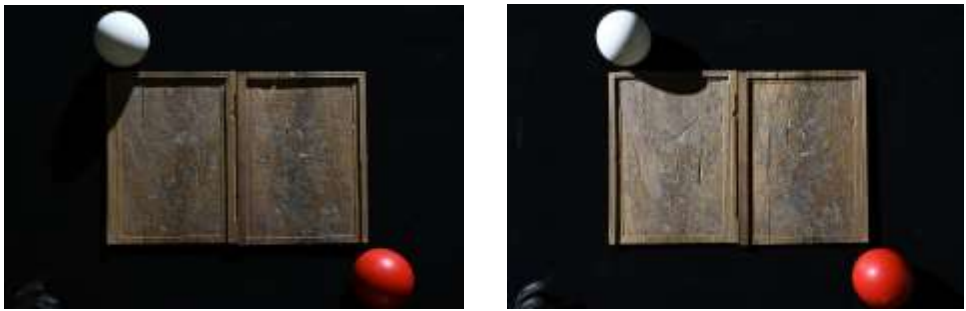

---

[23] Ingénieur au Ministère de la Culture, spécialiste des systèmes hydrauliques antiques.
[24] Dechezleprêtre s.d.

Figure 4 - La stéréophotométrie, appliquée à des images de la tablette zodiacale prises sous la même pose mais sous différents éclairages, fournit les résultats de la figure 5.

La figure 5 montre les deux résultats fournis par cette technique. Outre le champ de normales[25], représenté en RVB (rouge, vert, bleu), qui caractérise le relief, la couleur intrinsèque de la scène, ou albédo, est également estimée, ce qui constitue un avantage considérable lorsque l'objet est à la fois peint et gravé. Par ailleurs, l'agrandissement de la figure 6 montre que cette technique de numérisation 3D se distingue également par la résolution des estimations, qui est celle du capteur, en l'occurrence environ 25 microns.

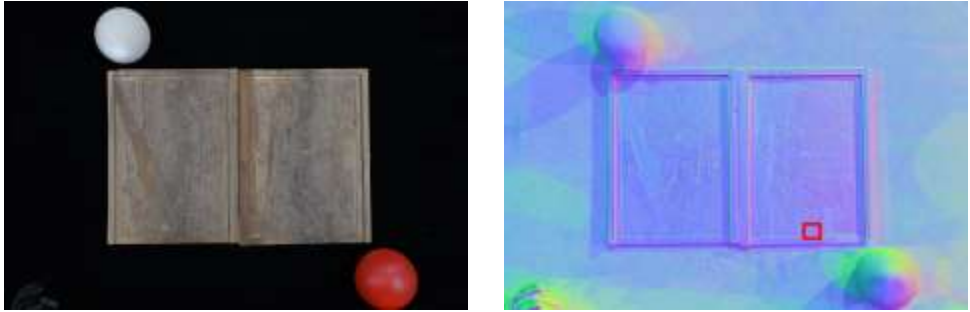

Figure 5 - Résultats de la stéréophotométrie appliquée à une dizaine d'images de la tablette zodiacale telles que celles de la figure 4 : à gauche, albédo (couleur intrinsèque) ; à droite, champ de normales en représentation RVB.

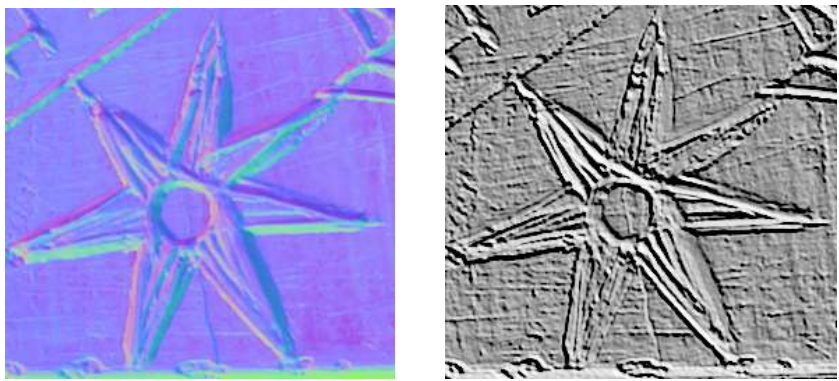

Figure 6 – Agrandissement de la zone encadrée en rouge sur le champ de normales de la figure 5 : à gauche , champ de normales ; à droite, rééclairage du champ de normales. La résolution est d'environ 25 microns.

**Utilisation de la stéréophotométrie pour l'aide à la lecture d'une tablette en bois**

La tablette MAN 2919 a été découverte à Mayence dans un terrain d'où est sortie la caliga MAN 2257. Elle a été achetée en 1865 auprès d'un antiquaire. Cette tablette en sapin était à l'origine enduite de cire. Le texte était gravé à l'aide de la pointe d'un stylet. Le texte pouvait être effacé par un simple grattage de la cire. Compte tenu de la nature périssable des matériaux et de l'usage donné à ces objets, on ne sait pas ce qui était inscrit. Les fantômes des écritures sont peut-être encore perceptibles parmi les traces incisées dans le bois tendre.

---

[25] Le champ de normales donne l'orientation locale de la surface en chaque pixel d'une image.

Parallèlement, l'observation dendrochronologique faite par François Blondel en 2023[26] livre la date de 35 av. n.è. pour le dernier cerne. Cette datation est précoce par rapport à celle qu'on donne habituellement des camps romains de Mayence. Certains indices plaident en effet pour un début d'occupation en 17 av. n.è.. La présence de troupes permanentes, sous la forme de deux légions, n'est cependant envisageable qu'à partir de 17 ap. n.è. avec l'établissement d'un camp de 37 ha. La datation de la tablette interroge la durée de vie de ces instruments d'écriture qu'on imagine aisément « consommables » et surtout les modalités de production de la filière bois et l'éventualité d'un réemploi des matériaux.

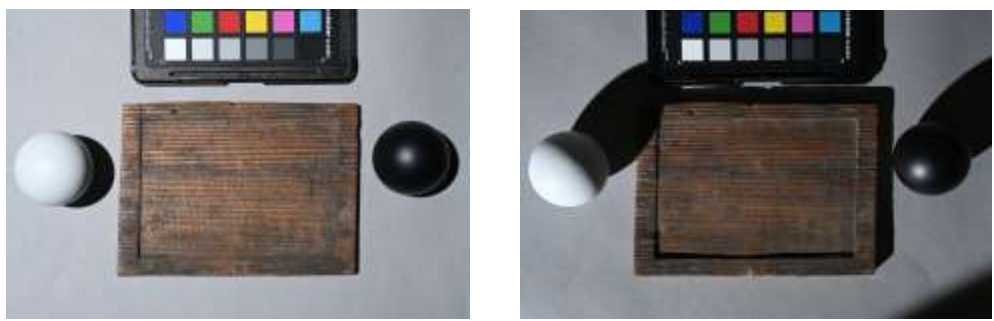

Figure 7 - Deux images de stéréophotométrie (parmi une dizaine) de la tablette de Mayence.

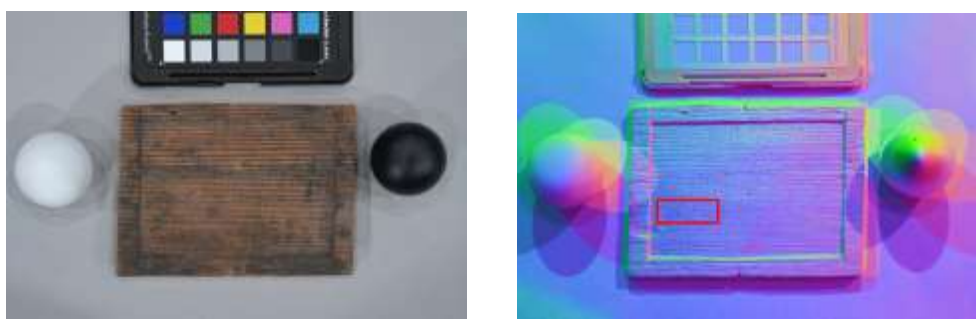

Figure 8 - Résultats de la stéréophotométrie appliquée à la tablette de Mayence (cf. figure 7) : albédo et champ de normales.

Par intégration du champ de normales estimé par stéréophotométrie, il est possible de calculer une carte de profondeur. C'est ce qui a été fait sur la partie encadrée en rouge sur le champ de normales de la figure 8, dont le résultat est montré sur la figure 9. La présence de caractères résultant de traces d'écriture pourra ainsi être mise en évidence à partir des observations des archéologues.

---

[26] Etude dendrochronologie inédite par F. Blondel.

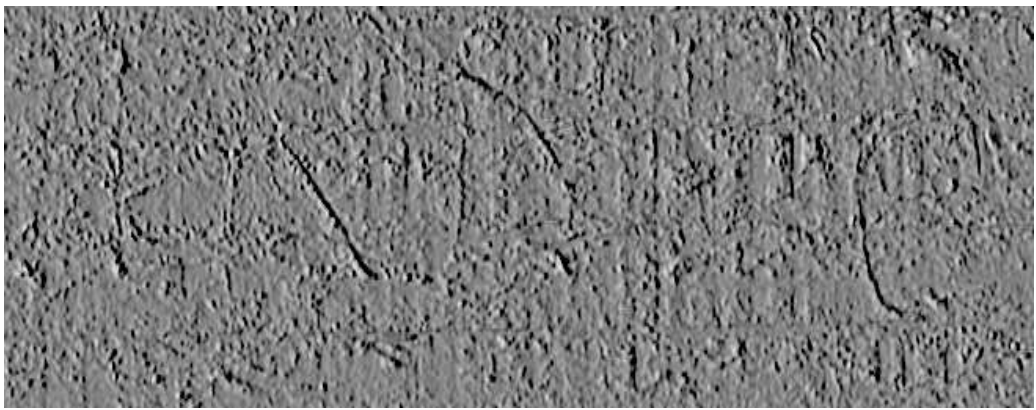

Figure 9 - Relief obtenu en intégrant les normales de la zone encadrée en rouge sur le champ de normales de la figure 8.

**Recours à la stéréophotométrie pour la lecture des gravures sur un coquillage du Mas d'Azil**

De 1887 à 1894, Édouard Piette (1827-1906), un pionnier de la préhistoire, fouille la grotte du Mas d'Azil sur la commune homonyme, en Ariège[27]. C'est la quatrième cavité pyrénéenne qu'il explore, après les grottes de Gourdan (Haute-Garonne), Lortet (Hautes-Pyrénées) et Espalungue-Arudy (Pyrénées-Atlantiques), dans les années 1870, et avant la grotte du Pape à Brassempouy (Landes), de 1894 à 1897. La grotte du Mas d'Azil est tellement vaste et riche que le mobilier archéologique qui y est découvert constitue la moitié de la collection Piette[28]. En 1904, le préhistorien cède cette dernière, longtemps conservée dans sa maison ardennaise, au musée des Antiquités nationales, devenu musée d'Archéologie nationale, qui est visible aujourd'hui dans la salle Piette, dont la présentation muséographique ancienne a été restaurée il y a une quinzaine d'années[29].

Le mobilier archéologique mis au jour dans les couches magdaléniennes de la grotte du Mas d'Azil comporte de nombreux coquillages perforés[30]. Ces éléments de parure pouvaient être portés en pendeloque ou cousus en applique sur les vêtements de peau. Quelques coquillages ont même été façonnés et gravés de motifs plus ou moins organisés. Un fragment d'un grand coquillage bivalve (MAN 47569), appartenant à l'espèce *Pecten maximus*, plus connue sous le nom de coquille Saint-Jacques, montre un reste de perforation et, sur sa face interne lisse, un décor gravé exceptionnellement figuratif. Il faut signaler ici que le *Pecten maximus* vit dans l'océan Atlantique et se trouve donc, dans la grotte du Mas d'Azil, à plusieurs centaines de kilomètres de son milieu d'origine, après un long voyage ou quelques échanges.

Les incisions, probablement réalisées avec la pointe d'un burin de silex taillé, sont d'une finesse telle qu'il faut un éclairage rasant pour les déchiffrer. Il s'agit d'une tête animale, représentée de profil droit. La forme des oreilles et des cornes, ainsi que le contour arrondi du museau, font penser à un bison. Le détail de l'œil et du pelage, notamment au niveau de la barbe et du chanfrein, sont caractéristiques du naturalisme magdalénien. La tête du bovidé est traversée de lignes gravées dans le sens des rainures de la coquille, ce qui rend sa lecture encore plus malaisée.

---

[27] Piette 1907.
[28] Thiault 1996.
[29] Schwab 2008.
[30] Fischer 1896.

Un relevé complet de ce coquillage et un relevé simplifié de son décor gravé, sans les lignes transversales, ont été publiés en 1896 par Henri Fischer, malacologue reconnu et neveu d'Édouard Piette, dans un article de la revue *L'Anthropologie*. Un cliché figure dans le catalogue de la collection Piette, rédigé par Marthe Chollot[31] et édité par les musées nationaux en 1964. La gravure n'est guère visible sur la photographie, mettant en évidence combien cette pièce est difficile à appréhender.

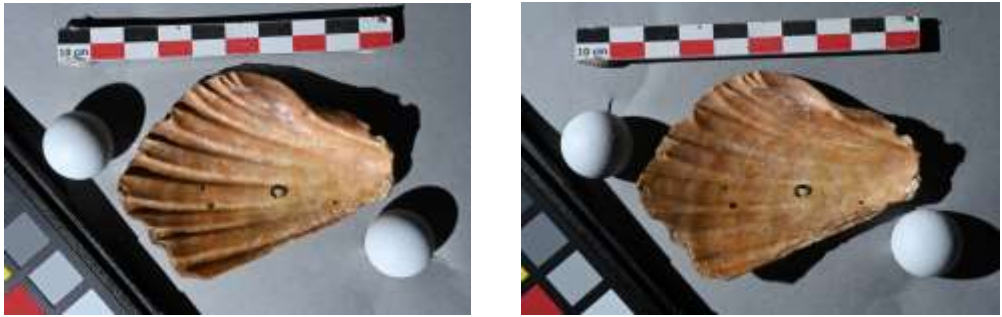

Figure 10 - Deux images de stéréophotométrie (parmi une dizaine) du coquillage gravé.

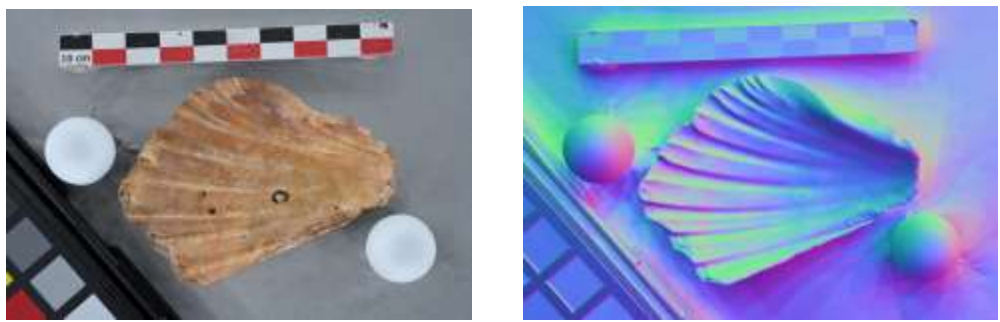

Figure 11 - Résultats de la stéréophotométrie appliquée au coquillage gravé (cf. figure 10) : albédo et champ de normales.

Le résultat de la figure 12 permettra à la fois de mettre en évidence le motif, et surtout d'appréhender la chronologie relative des tracés. Dans l'agrandissement, il est aisé d'observer quel tracé recoupe un autre et d'en déduire sa postériorité. Par la multiplication de ces observations au niveau des croisements, l'archéologue reconstruit les étapes de la mise en place des motifs.

---

[31] Chollot 1964.

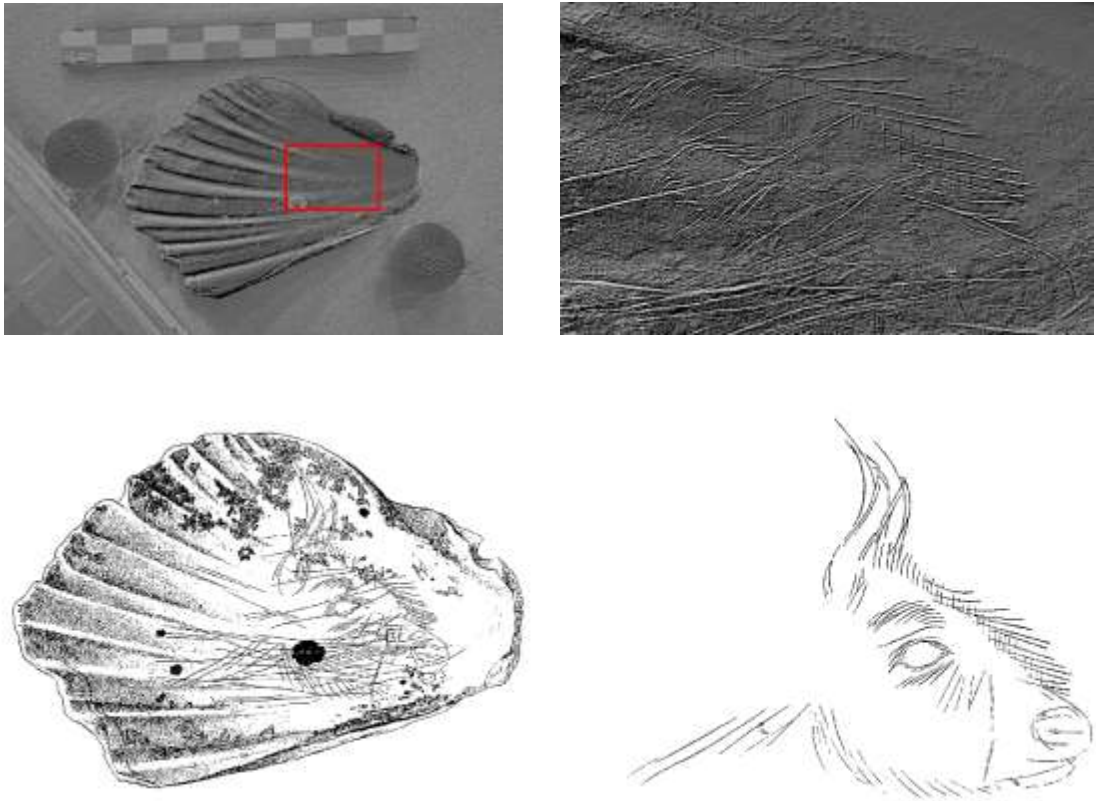

Figure 12 - Le coquillage gravé de la grotte du Mas d'Azil, en Ariège (MAN 47569), date du Magdalénien, entre –20500 et –13000 ans environ. D'une longueur de 8,1 cm, d'une largeur de 5,5 cm et d'une épaisseur de 0,4 cm, il figure une tête de bovidé. En rééclairant le champ de normales et en agrandissant la zone encadrée en rouge, ce motif apparaît clairement. Relevé de la face interne gravée et relevé simplifié de la gravure, publiés par H. Fischer en 1896.

**Utilisation de données de RTI en stéréophotométrie**

La RTI (*reflectance transformation imaging*) est une technique particulièrement appréciée par les archéologues, qui permet de faire apparaître le dessin des gravures les plus ténues en jouant sur l'éclairage. En effet, ce dernier, lorsqu'il est rasant, provoque l'apparition d'ombres très fines. Si le nombre de directions d'éclairage est suffisamment important au moment de la captation des données et couvre bien l'ensemble des directions possibles, la simulation d'une nouvelle direction d'éclairage, qui est obtenue par interpolation, est saisissante de réalisme, y compris pour des matériaux translucides ou transparents.

La RTI n'est pas à proprement parler une technique de reconstruction 3D. Or, étant donné que le protocole d'acquisition des données est très similaire à celui de la stéréophotométrie, puisque la pose de l'appareil photographique doit être fixe, il est naturel d'exploiter ces données par stéréophotométrie. La redondance des données, lorsque la RTI utilise une centaine d'images là où la stéréophotométrie peut se contenter d'un minimum de trois éclairages, peut être exploitée pour rendre l'estimation des normales plus robuste.

L'utilisation d'un dôme de RTI, dont les éclairages peuvent être étalonnés, simplifie grandement l'acquisition, puisqu'il devient inutile de positionner des sphères au voisinage de la scène 3D. Néanmoins, la RTI est plutôt prévue pour analyser des scènes quasi-planes, pour lesquelles une seule pose suffit à caractériser le relief par le biais d'une carte de profondeur. Nous avons choisi de mettre en œuvre cette technique sur une mosaïque du MAN.

**Étude d'une mosaïque à partir de données de RTI**

La mosaïque des Saisons de Saint-Romain-en-Gal (MAN 83116) date du début du IIIe siècle ap. J.-C. Cette mosaïque, dont plus du tiers est manquante (4,48 m sur 8,86 m ont été conservés), ornait le sol d'une grande demeure suburbaine de Saint-Romain-en-Gal, un des quartiers de Vienne dans l'Antiquité. C'était l'une des villes les plus belles et les plus prospères de la Gaule. Des ateliers de mosaïstes installés à Vienne satisfaisaient la demande d'une clientèle aisée.

Les activités agricoles et les fêtes des quatre saisons sont représentées en 40 tableaux de 59 cm chacun, dont 27 subsistent, insérés dans une riche tresse décorative. Ce thème n'est pas souvent utilisé par les mosaïstes en général, et par les ateliers de Vienne en particulier, qui préfèrent représenter des scènes mythologiques ou des natures mortes. Ils se sont sans doute inspirés d'un ou plusieurs modèles romains, valables pour l'ensemble des pays méditerranéens, aussi ne faut-il pas considérer cette mosaïque comme un « reportage » sur les pratiques agricoles gallo-romaines.

Les différents tableaux s'articulent autour des quatre tableaux centraux où figurent des personnifications des Saisons : l'Hiver est une femme emmitouflée sur un sanglier, le Printemps un Amour nu sur un taureau, l'Été un Amour nu sur un lion, l'Automne un Amour nu sur un tigre.

Un vaste projet de restauration à l'atelier de restauration des mosaïques et des enduits peints de Saint-Romain-en-Gal a donné lieu à une numérisation 3D par photogrammétrie de la mosaïque assemblée, ainsi qu'à la prise de vues de très haute résolution. Ces supports, combinés à des outils de géomatique, offrent d'intéressantes perspectives d'études et de recherche (cartographie des restaurations, automatisation du dénombrement des tesselles, étude d'usure des surfaces, etc.) et de médiation, dans le cadre d'un projet d'exposition.

La figure 13 montre le dôme de RTI en situation de prise de vues, directement posé sur la mosaïque. Il dispose de 105 LEDs réparties sur une surface hémisphérique. Étant donné que cette structure rigide est fixée sur l'objectif de l'appareil photographique, les éclairages peuvent être étalonnés une fois pour toutes, ce qui allège le protocole d'acquisition.

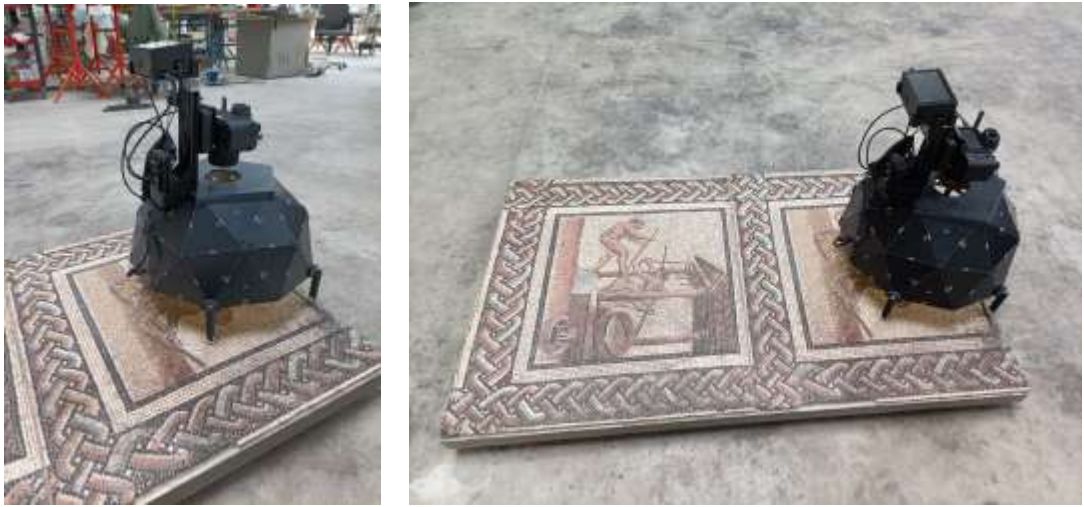

Figure 13 - Deux photographies du dôme de RTI posé sur la mosaïque de Saint-Romain-en-Gal.

La figure 14 montre deux des 105 images de RTI de la mosaïque, que nous utilisons comme données de stéréophotométrie. Les résultats sont ceux de la figure 15. L'utilisation d'un dôme de RTI étalonné permet d'éviter les artéfacts dus aux ombres des sphères, qui sont nettement visibles sur les champs de normales des figures 5 et 8. En outre, la redondance des données de RTI, en comparaison des données usuelles de stéréophotométrie, ouvre plusieurs perspectives : une estimation plus robuste des normales et de l'albédo, l'estimation d'autres propriétés des matériaux que leur simple couleur, voire l'estimation de l'état de surface des tesselles.

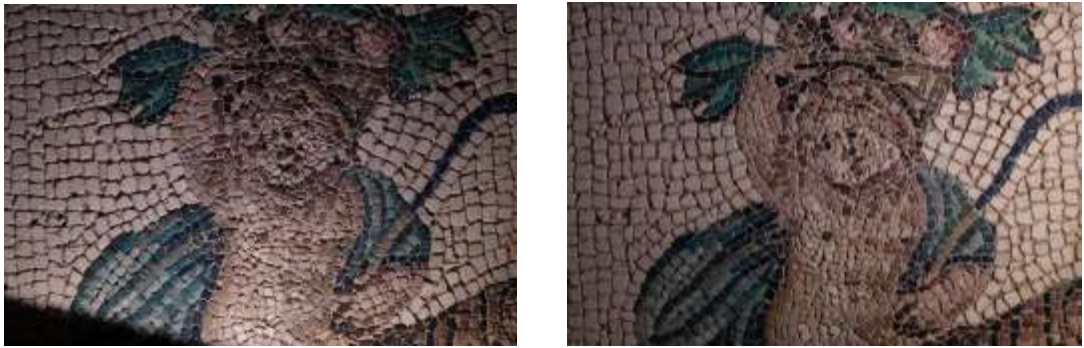

Figure 14 - Deux photographies de la mosaïque de Saint-Romain-en-Gal (parmi une centaine) prises avec un dôme de RTI.

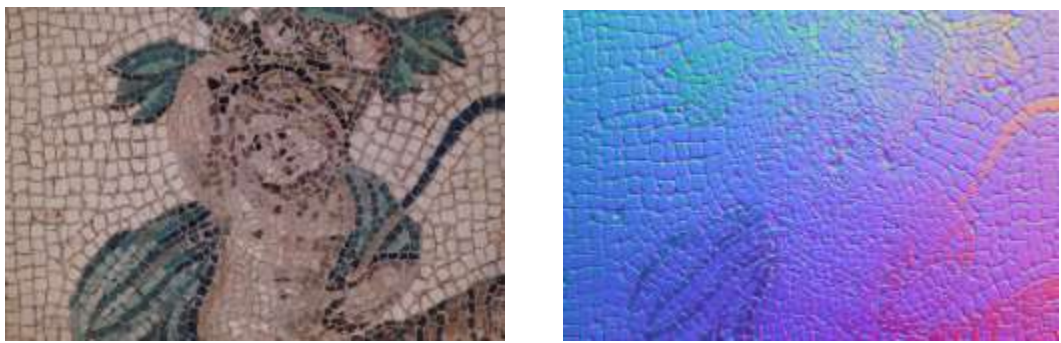

Figure 15 - Résultats de la stéréophotométrie appliquée aux images de RTI (cf. figure 14) : albédo et champ de normales.

**Perspectives de l'approche stéréophotométrique**

De nombreuses perspectives pourraient compléter cette étude méthodologique, en particulier l'usage des données de RTI, très redondantes, réplicables grâce à l'utilisation d'un dôme étalonné en amont, qui doivent nous permettre d'aller plus loin dans l'étude des objets, en caractérisant par exemple leur état de surface.

Une des limites de la stéréophotométrie est due aux conditions opératoires qui nécessitent de contrôler l'éclairage, ce qui empêche, en pratique, son utilisation en extérieur. Saurons-nous adapter cette méthode pour numériser les gravures des dolmens dans les fossés du château de Saint-Germain-en-Laye ? La comparaison entre différentes copies numériques permettrait, par exemple, de suivre leur évolution temporelle, pour peu que les numérisations puissent être recalées avec précision dans un repère commun.

Enfin, l'utilisation conjointe de plusieurs poses et de plusieurs éclairages doit nous permettre de répondre mieux encore aux besoins des archéologues.

**Conclusion**

Dans le domaine du patrimoine, l'obtention de modèles de référence est par définition impossible, puisque les objets ont été façonnés par des méthodes non numériques. Contrairement à certaines productions actuelles qui disposent d'un modèle provenant de la conception assistée par ordinateur, le jumeau numérique ne peut pas être comparé à cette référence. Il est envisageable de produire un modèle 3D avec différents outils sans avoir la certitude d'approcher la réalité de l'objet, que ce soit sa géométrie, sa couleur ou les propriétés de ses matériaux.

Nous avons tenté de montrer dans cet article qu'il n'existait pas de méthode de production de modèles 3D adaptée à tous les types d'objets patrimoniaux. Bien sûr, nous essayons toujours de tendre vers la réplique de la réalité physique du vestige archéologique. Nous pouvons éventuellement limiter l'écart entre le vestige et sa réplique par synthèse d'images, en comparant les images simulées aux images d'origine. Ceci nous permet d'insister sur la mission des musées pour la conservation des œuvres, alors même que le jumeau numérique devient une archive, certes limitée, de l'objet, et ouvre le débat sur l'impact et le poids des données produites (gestion / archivage de la donnée / métadonnées), notamment vis-à-vis de son usage.

**Bibliographie**